\documentclass[10pt,journal]{IEEEtranTCOM}

\usepackage[english]{babel}
\usepackage[usenames]{color}
\usepackage[cp1250]{inputenc}
\usepackage{amsfonts}
\usepackage{amsthm}
\usepackage{graphicx}
\usepackage{epsfig}
\usepackage{mathrsfs}
\usepackage{amsmath}
\usepackage{algorithm}
\usepackage{algorithmic}
\usepackage{hyperref}

\theoremstyle{plain}

\begin{document}
\newcommand{\bea}{\begin{eqnarray}}
\newcommand{\eea}{\end{eqnarray}}
\newcommand{\be}{\begin{equation}}
\newcommand{\ee}{\end{equation}}
\newcommand{\beas}{\begin{eqnarray*}}
\newcommand{\eeas}{\end{eqnarray*}}
\newcommand{\bs}{\backslash}
\newcommand{\bc}{\begin{center}}
\newcommand{\ec}{\end{center}}
\def\SC {\mathscr{C}}
\def\ind {\perp\!\!\!\perp}

\title{Extracting individual variable information\\for their decoupling, direct mutual information,\\and multi-feature Granger causality}
\author{\IEEEauthorblockN{Jarek Duda}\\
\IEEEauthorblockA{Jagiellonian University,
Golebia 24, 31-007 Krakow, Poland,
Email: \emph{dudajar@gmail.com}}}
\maketitle

\begin{abstract}
Working with multiple variables they usually contain difficult to control complex dependencies. This article proposes extraction of their individual information, e.g.  $\overline{X|Y}$ as random variable containing information from $X$, but with removed information about $Y$, by using $(x,y) \leftrightarrow (\bar{x}=\textrm{CDF}_{X|Y=y}(x),y)$ reversible normalization. One application can be decoupling of individual information of variables: reversibly transform $(X_1,\ldots,X_n)\leftrightarrow(\tilde{X}_1,\ldots \tilde{X}_n)$ together containing the same information, but being independent: $\forall_{i\neq j} \tilde{X}_i\ind \tilde{X}_j, \tilde{X}_i\ind X_j$. It requires detailed models of complex conditional probability distributions - it is generally a difficult task, but here can be done through multiple dependency reducing iterations, using imperfect methods (here HCR: Hierarchical Correlation Reconstruction). It could be also used for direct mutual information - evaluating direct information transfer: without use of intermediate variables. For causality direction there is discussed multi-feature Granger causality, e.g. to trace various types of individual information transfers between such decoupled variables, including propagation time (delay).
\end{abstract}
\section{Introduction}
While basic motivation of machine learning is building "black box" predictors trained on data, in recent years there is a growing emphasis to try to make them also explain the used mechanisms, what requires better understanding of relations deeply hidden in the data, like building Bayesian networks~\cite{bayes} of hidden causality relations between variables.

While ideally we would like to automatically discover such dependencies from the data, it is relatively difficult. It seems natural to try to use information theoretic tools like mutual information evaluating information transfer in bits, however, such information transfer is not necessarily direct - might go through some other intermediate nodes/variables.

\begin{figure}[t!]
    \centering
        \includegraphics{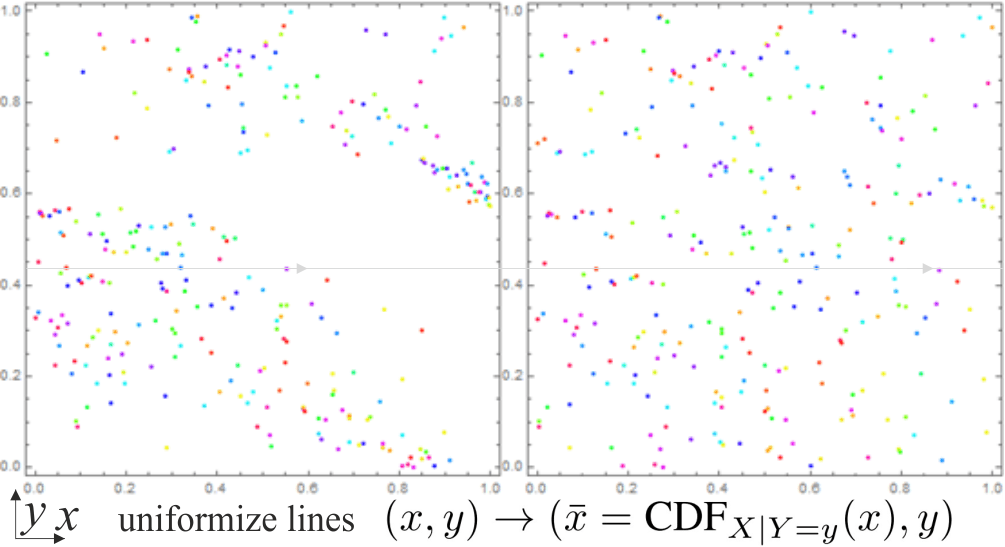}
        \caption{Example of extraction of individual information of $X$ variable, by removing information already available in $Y$ variable, both initially normalized to nearly $U[0,1]$ (marginal) distributions. There was modeled their joint distribution $\rho(x,y)$ as polynomial (HCR), substituted various $y$ and by integration calculated $\bar{x}=\textrm{CDF}_{X|Y=y}(x)=\int_0^x \rho(z,y) dz/\int_0^1 \rho(z,y) dz$, uniformizing each "$y=\textrm{const}$" horizontal line by shifting left/right the shown points of dataset (marked with random colors, the same for both diagrams). Such new $\bar{X}$ variable is from nearly $U[0,1]$ distribution for all $y$ - no longer providing information about $Y$ variable, but still containing individual information of $X$ as we can reverse this (bijective) transformation $(\bar{X},Y)\leftrightarrow (X,Y)$. Notice that this transformation modifies entropy as it contains normalization (avoidable?). }
        \label{intr}
\end{figure}

This article proposes tools to literally remove information already available in some other variables like in Fig. \ref{intr}, allowing to extract individual information of each variable, decouple them into independent variables together containing the same information. For example to evaluate direct information transfer, or enhance Granger-type causality to focus on relations between decoupled individual variables. We could also combine decoupling with some interpretable model to distinguish contributions from individual variables.

Another type of applications might be algorithmic fairness, ethical machine learning: often we would like to remove some information (like sex, age, ethnicity) from the actually used variables to avoid bias~\cite{ethical} - the proposed approach could literally extract the only information the machine is allowed to exploit. 

This is initial version of article, to be extended in the future especially by real life applications.

\section{Individual information extraction}
This main Section discusses the basic proposed tools.

\subsection{Variable normalization to quantiles}
Having a random variable $X$, a basic normalization is subtraction of mean and division by standard deviation. It is natural for Gaussian variables, however, e.g. for Student's t-distribution such variable would still have some shape parameter, and heavy tails. Therefore, e.g. in copula theory~\cite{copula}, there is popular stronger normalization to (approximately in practice) uniform distribution on $[0,1]$ (used in this article), referred as $U[0,1]$:
\be \bar{X}:=\textrm{CDF}_X(X)\qquad \textrm{for dataset:}\quad \bar{x}=\textrm{CDF}_X(x)\ee
where $\textrm{CDF}_X(x)=\int_{-\infty}^x \rho_X(y)dy$ is cumulative distribution function of $X$, making such normalized variable $\bar{X}\sim U[0,1]$ corresponding to quantiles of the original variable, e.g. median to $\bar{x}=1/2$.

In practice we rarely really know $\textrm{CDF}_X$, what is resolved in two basic ways: assume some parametric distribution and e.g. estimate its parameters for a given dataset, or use empirical distribution function: sort the dataset values and assign to each value its position in ordering rescaled to $[0,1]$.

\subsection{Extraction: $\overline{X|Y}$ removing $Y$ information from $X$}
The basic discussed tool is perfoming analogous normalization, but using CDF conditioned on a second (or multiple) variable like in Fig. \ref{intr} - \emph{extracting} individual information of $X$ in $(X,Y)$ pair of variables:
\be (x,y) \to \left(\bar{x}=\textrm{CDF}_{X|Y=y}(x),y\right) \ee
\be \overline{X|Y}:=\textrm{CDF}_{X|Y}(X)\ee

As in this figure, it is performing such normalization to $U[0,1]$ separately for each "$y=\textrm{const}$" line of $(x,y)$ plot by shifting inside lines: $\textrm{Pr}(\textrm{CDF}_{X|Y=y}(x)\leq\bar{x})=\bar{x}$. Hence such $\bar{X}=\overline{X|Y}$ variable becomes independent from $Y$: $\bar{X}\ind Y$.

 As $\textrm{CDF}^{-1}(\textrm{CDF}(x))=x$ for all possible values of $X$, knowing this $\textrm{CDF}$ we can reverse this operation:
\be (\bar{x},y) \to \left(x=\textrm{CDF}^{-1}_{X|Y=y}(\bar{x}),y\right) \ee

Therefore, we can see $(X,Y)$ and $(\overline{X|Y}|Y)$ as equivalent $(\equiv)$: transformable in reversible way, hence considered together (jointly) they carry the same information. 

As modelling of conditional CDFs is quite difficult, we could also calculate it by \emph{iterations}, e.g. two: $\overline{\overline{X|Y}|Y}$ or more. Initial experimental tests with HCR show they indeed improve independence, still maintaining individual information of $X$.


While the above is natural for continuous variables, transformation $(x,y) \leftrightarrow (\bar{x}=\textrm{CDF}_{X|Y=y}(x),y)$ has no problem if $Y$ is not continuous (discrete or mixed) - e.g. for discrete just use separate $\textrm{CDF}_{X|Y=y}$ for each discrete $y$. However, for discrete $X$ such $\bar{x}=\textrm{CDF}_{X|Y=y}(x)$ might have a larger number of discrete values for discrete $Y$, or even can become continuous for non-discrete $Y$.

\subsection{Decoupling variables and symmetrization}
Applying the discussed reversible transformation twice:
\be (X,Y) \equiv (\tilde{X}=\overline{X|Y},Y) \equiv (\tilde{X},\tilde{Y}=\overline{Y|\tilde{X}})  \ee
or generally $n$ times for $n$ succeeding variables of $(X_1,\ldots,X_n)$, we get reversible transformation to $(\tilde{X}_1,\ldots \tilde{X}_n)$ together (jointly) containing the same information, but now being independent: $\forall_{i\neq j} \tilde{X}_i\ind X_j, \tilde{X}_i\ind \tilde{X}_j$. Hence such extraction for each variable of remaining available information can be thought of as \emph{decoupling} of their individual information.

As practical models of conditional distributions are rather far from perfect, it might be worth to perform it multiple times (\emph{iterative decoupling}), each time reducing dependencies, not necessarily using natural order, maybe directly using information from earlier iterations. There were performed some basic tests using HCR, confirming that such iterations reduce dependencies maintaining individual information, optimization of this process will require further work.

Such iterative decoupling depends on arbitrary choice of order of information extractions. It generally does not seem an issue, but there could be also considered symmetrized version, e.g. performing many iterations of steps reduced to nearly infinitesimal, or maybe using values from single iteration at cost of reversibility, e.g. $\forall_i\ \tilde{X}_i =\overline{X_i|X_1,\ldots,X_{i-1},X_{i+1},\ldots,X_n}$.

\subsection{Direct mutual information}
In popular Bayesian networks~\cite{bayes}, for a dataset on  $(X_1,\ldots,X_n)$ variables, we would like to automatically build DAG (directed acyclic graph) showing direct causalities between variables. From information theory perspective, the closest evaluation of two variables being connected (without causality direction) seems mutual information, especially if  normalizing these variables (e.g. not to depend on arbitrary scales).

While such mutual information ($I(X;Y)=H(X)-H(X|Y)$) can be imagined  as the number of bits of information transmitted between these two variables (in any direction), it could be transmitted through some other intermediate variables - to propose an edge in Bayesian network, we should estimate how many bits are transmitted directly. 

Assuming we know all such possible intermediate nodes denoted as $Z$ (e.g. $X=X_1, Y=X_2, Z=(X_3,\ldots,X_n)$), we could first extract their individual information (also normalize) to $\overline{X|Z}$ and $\overline{Y|Z}$, and then calculate their mutual information - evaluating their normalized direct dependence:
\be \textrm{\emph{direct mutual information}:}\ I_d(X;Y|Z)= I\left(\overline{X|Z};\overline{Y|Z}\right)\ee

It is strongly related to \emph{conditional mutual information}~\cite{cond}:
$$I(X;Y|Z)=H(X|Z)-H(X|Y,Z)=I(X;Y,Z)-I(X;Z)$$
with included normalization and being more practical to estimate thanks to the mentioned iterative extraction. Direct estimation of $I(X;Y|Z)$ for $Z$ being a large number of variables is technically difficult, especially that it is a difference of two close noisy values.

\section{HCR (Hierarchical Correlation Reconstruction) to estimate CDFs}
In HCR~\cite{HCR} family of methods\footnote{Introduction to HCR: \url{https://community.wolfram.com/groups/-/m/t/3017754}} to work with probability distributions, we usually first normalize variables to nearly uniform on $[0,1]$, then model (joint) density (PDF) as a linear combination in some orthonormal basis $(f_i)_{i=0..m}$ (usually polynomials): $\rho(x)=\sum_{i=0}^m a_i f_i(x)$. These $a_i$ are moments for polynomial basis, and thanks to orthonormality their MSE estimation turns out just mean of $f_i(x)$ over $x$ from dataset. Such density as a linear combination can get below 0, hence there is required further e.g. $\max(\rho,0.1)/Z$ calibration as the final density model, where $Z$ is normalization to integrate to 1.

To estimate conditional CDFs, we can analogously use it to model joint density, substitute to it the variables we condition on, normalize obtained density for the remaining variables, and finally integrate to get conditional CDF.

However, such \emph{joint distribution approach} is very difficult to make practical for large numbers of variables. Therefore, especially for this purpose there was proposed alternative approach: to \emph{directly predict conditional distributions}, successful for various types of data (\cite{hcr1,hcr2,hcr3,hcr4}).

Specifically, to predict conditional density $X|Y$, where $Y$ may contain multiple variables, we represent conditional density as 
\be \rho(x|y)=1+\sum_{i=1}^m f_i(x) a_i(y) \label{dens}\ee
and find/train estimator for $a_i(y)$ minimizing MSE (mean squared error): averaged  $\|f_i(x)-a_i(y)\|_2^2$ over all $(x,y)$ from dataset, separately for each $i$-th predicted moment. 

Such $a_i(y)$ estimator of $f_i(x)$ can be just a linear regression from some features of (variables of) $Y$, can be also more sophisticated e.g. neural network - instead of standard direct prediction of value $x$ from $y$, here we separately MSE predict some its $i$-th moments: $f_i(x)$ from $y$, and finally combine these moment predictions into predicted density (\ref{dens}).

Both approaches were initially tested for discussed extraction, some results and discussion is planned for future versions of this article.

\section{Multi-feature Granger causality}
Determination of causality directions e.g. for the found edges is very difficult from data alone, e.g. $Y\sim g(X)+$ noise for some function $g$ would suggest $X$ causes $Y$. 

Another basic tool is obvious observation that for two related events in different times, it is rather the earlier one causing the later one. However, alternatively there could be an even earlier common cause. 

Such imperfect type of causality is referred as Granger's~\cite{granger}: $(Y_t)$ time series "G-causes" $(X_t)$, if past values of $(Y_t)$ contain information allowing to improve prediction of $(X_t)$ above and beyond the information contained in past values of $(X_t)$ alone.

Usually such analysis is realized by linear regression of $(X_t)$ from its past values, and separately from past values of both $(X_t)$ and $(Y_t)$ - testing if it is essentially better.

Recently proposed \emph{multi-feature Granger causality}~\cite{EEG}, enhances the above true/false evaluation in two ways:
\begin{enumerate}
  \item Analyze \textbf{delay $\Delta t$ dependence}. A basic way is to first calculate residues: $r_t=x_t -$ "its prediction from its past values". Then calculate correlations among all $(r_t,y_{t-\Delta t})_t$ pairs, separately for various $\Delta t$ - delay maximizing correlation can be interpreted as the propagation time. It might be also worth calculating Fourier transform of such $\Delta t$ dependence to find characteristic frequencies.
  \item \textbf{Multi-feature} - above correlation is only one of statistical dependencies calculated from joint distribution. We could get different types by analyzing  $\Delta t$ dependence of the entire joint distribution - discussed further.
\end{enumerate}
For multi-feature, instead of residue as subtracted predicted values, there was proposed (elaborated here) extraction of individual information - as residue use $R_t=\overline{X_t|X_{t-1}X_{t-2}\ldots}$ - with removed information from its past, predicting not only value, but its entire probability distribution (e.g. value + variance).

Now we would like to evaluate joint distribution between such (both normalized) $(R_t,Y_{t-\Delta t})_t$ separately for various $\Delta t$. We could use e.g. mutual information to evaluate the amount of transmitted bits of information. But alternatively we could extract multiple features from these joint distributions, the further ones might indicate some subtle but valuable dependencies. With HCR it is natural to estimate this $\rho_{\Delta t}(r,y)$ joint density of $(R_t, Y_{t-\Delta t})$ as a linear combination (e.g. polynomial): $a_{jk}(\Delta t)$ coefficients, which can be further reduced to a few dominant contributions $a_i(\Delta t)$ with PCA dimensionality reduction:
\be \rho_{\Delta t}(r,y)\stackrel{\text{HCR}}{\approx} \sum_{j,k=0}^m f_j(y) f_k(z)\, a_{jk}(\Delta t)
   \stackrel{\text{PCA}}{\approx}1+ \sum_{i=1}^r   f_{v_i}(r,y)\, a_i(\Delta t) \ee

With proposed here extraction we can enhance it to get individual causality between multiple variables. For this purpose, having $(X^1_t,\ldots, X^n_t)$ time series, first decouple them: convert to $(\tilde{X}^1_t,\ldots, \tilde{X}^n_t)\equiv (X^1_t,\ldots, X^n_t)$ with $\forall_{i\neq j} \tilde{X}_i\ind X_j$, and for decoupled test the discussed multi-feature Granger causality.

\section{Conclusions and further work}
There was proposed extraction of individual information from variables - removing information contained in the remaining ones, e.g. to literally decouple them: transform to independent, but together containing the same information. Or to estimate direct information transfer between pairs of variables: removing transfer through some intermediate nodes, or multi-feature Granger causality to evaluate multiple individual directional relations and their delays (propagation time).

This is initial version of the article, to be expanded in the future e.g. by:
\begin{itemize}
  \item Development of applications, tests, comparison with alternatives. 
  \item Search for new applications for the proposed tools, like bias removal~\cite{ethical} for ethical ML, or combination of decoupling with some interpretable model to extract individual contributions.
  \item Optimize tools: details of HCR, regularization e.g. "lasso", ordering especially for iterations, symmetrization, adding neural networks, also search for different tools than HCR.
  \item The discussed extraction contains normalization which modifies information theoretic evaluations - the question is if it could be avoided or compensated (e.g. rescale $\overline{X|Y}$ variable to make $H(\overline{X|Y})=H(X|Y)$?), also relate with other especially information theoretic tools (e.g. direct vs conditional mutual information).
\end{itemize}

\bibliographystyle{IEEEtran}
\bibliography{cites}
\end{document}